\documentclass[conference,10pt]{IEEEtran}
\IEEEoverridecommandlockouts
\usepackage[colorlinks=false]{hyperref}
\usepackage{amsmath,amssymb,amsfonts}
\usepackage{adjustbox}
\usepackage{algorithmic}
\usepackage{threeparttable}
\usepackage{array}
\usepackage{fontawesome}
\usepackage{graphicx}
\usepackage{multirow}
\usepackage{textcomp}
\usepackage{enumitem}
\usepackage[english]{babel}
\usepackage[utf8]{inputenc}
\usepackage[nolist]{acronym}
\usepackage{amsmath}											
\usepackage{amsfonts}											
\usepackage{blindtext} 											
\usepackage{booktabs}											
\usepackage{fancyvrb}
\usepackage{ifthen}												
\usepackage{filecontents} 										
\usepackage[T1]{fontenc} 
\usepackage{graphicx}
\usepackage[dvipsnames,table]{xcolor}
\usepackage{ifthen}												
\usepackage{listings}											
\usepackage{lipsum}
\usepackage{multirow}
\usepackage{pgfplots}
\usepackage{siunitx}
\usepackage{ragged2e}						
\usepackage{tabularx}
\usepackage{url}
\usepackage{tikz}
\usepackage{circuitikz}
\usepackage{comment}
\usepackage{footnote}
\usepackage{float}
\usepackage{listings}
\usepackage{setspace}
\usepackage{makecell, tabularx}
\usepackage{enumitem}

\usetikzlibrary{calc,trees,positioning,arrows,chains,shapes.geometric,%
	decorations.pathreplacing,decorations.pathmorphing,shapes,%
	matrix,shapes.symbols}

\usepackage[font=footnotesize,labelfont=bf]{subcaption}		
\usepackage{caption}
\usepackage[ruled,linesnumbered]{algorithm2e}
\SetAlCapNameFnt{\small}
\SetAlCapFnt{\small}
\SetAlgoNlRelativeSize{0}
\RestyleAlgo{ruled}
\SetKwComment{Comment}{/* }{ */}

\usepackage[style=ieee, backend=biber, natbib=true, maxbibnames=1, minbibnames=1]{biblatex}
\usepackage{fancyhdr}

\fancypagestyle{firstpage}{%
	\fancyhf{}%
	\fancyhead[C]{\normalsize To appear at the 38th SBC/SBMicro/IEEE Symposium on Integrated Circuits and Systems Design (SBCCI), \\ August 25-29, 2025, Manaus, BRAZIL}
	
	\fancyfoot[C]{\footnotesize \copyright2025 IEEE.  Personal use of this material is permitted.  Permission from IEEE must be obtained for all other uses, in any current or future media, including reprinting/republishing this material for advertising or promotional purposes, creating new collective works, for resale or redistribution to servers or lists, or reuse of any copyrighted component of this work in other works.}
}

\newboolean{blindreview}
\setboolean{blindreview}{false}
\DeclareRobustCommand{\IEEEauthorrefmark}[1]{\smash{\textsuperscript{\footnotesize #1}}}

\usepackage[left=1.57cm,right=1.57cm,top=2cm,bottom=2.5cm,letterpaper]{geometry}

\begin{document}	
	
\begin{acronym}[] 
	\acro{CEX}[CEX]{Counter Example}
	\acrodefplural{CEX}[CEXs]{Counter Examples}
	\acro{HITL}{Human-in-the-Loop}
	\acro{IC}[IC]{Integrated Circuit}
	\acrodefplural{ICs}[ICs]{Integrated Circuits}
	\acro{LLM}[LLM]{Large Language Model}
	\acrodefplural{LLM}[LLMs]{Large Language Models}
	\acro{MAS}{Multi-Agent System}
	\acro{RTL}{Register Transfer Level}
	\acro{SVA}[SVA]{SystemVerilog Assertion}
	\acrodefplural{SVA}[SVAs]{SystemVerilog Assertions}
	\acro{temp}{temperature}
	\acro{GenAI}{Generative AI}
\end{acronym}


\title{Hey AI, Generate Me a Hardware Code! \\ Agentic AI-based Hardware Design \& Verification}

\ifthenelse{\boolean{blindreview}}{}{
	\author{
		\IEEEauthorblockN{
			Deepak Narayan Gadde\IEEEauthorrefmark{1},
			Keerthan Kopparam Radhakrishna\IEEEauthorrefmark{1}, 
			Vaisakh Naduvodi Viswambharan\IEEEauthorrefmark{1},\\
			Aman Kumar\IEEEauthorrefmark{2},
			Djones Lettnin \IEEEauthorrefmark{3},
			Wolfgang Kunz\IEEEauthorrefmark{4},
			Sebastian Simon\IEEEauthorrefmark{1}}
		\IEEEauthorblockA{
			\IEEEauthorrefmark{1}Infineon Technologies Dresden GmbH \& Co. KG, Germany\\
			\IEEEauthorrefmark{2}Infineon Technologies India Pvt. Ltd., India\\
			\IEEEauthorrefmark{3}Infineon Technologies AG, Germany\\
			\IEEEauthorrefmark{4}Rheinland-Pf{\"a}lzische Technische Universit{\"a}t Kaiserslautern-Landau, Germany}
		\vspace*{-1cm}
		}
}

\maketitle

\thispagestyle{firstpage}

\begin{abstract}

Modern \acfp{IC} are becoming increasingly complex, and so is their development process. Hardware design verification entails a methodical and disciplined approach to the planning, development, execution, and sign-off of functionally correct hardware designs. This tedious process requires significant effort and time to ensure a bug-free tape-out. The field of Natural Language Processing has undergone a significant transformation with the advent of \acp{LLM}. These powerful models, often referred to as \ac{GenAI}, have revolutionized how machines understand and generate human language, enabling unprecedented advancements in a wide array of applications, including hardware design verification. This paper presents an agentic AI-based approach to hardware design verification, which empowers AI agents, in collaboration with \ac{HITL} intervention, to engage in a more dynamic, iterative, and self-reflective process, ultimately performing end-to-end hardware design and verification. This methodology is evaluated on five open-source designs, achieving over \SI{95}{\percent} coverage with reduced verification time while demonstrating superior performance, adaptability, and configurability.

\end{abstract}


\section{Introduction} \label{sec:introduction}
The swift progress in semiconductor industry has enabled the creation of intricate \ac{IC} designs with multiple functions on a single chip. However, this complexity has turned \ac{RTL} design and verification into significant bottlenecks in \ac{IC} development. \ac{RTL} design requires extensive manual expertise to ensure functional accuracy, optimal performance, and power efficiency whereas, verification stage alone can consume up to \SI{60}{\percent} of the total project time \cite{VerStudy}. Even with advancements in EDA tools, verification inefficiencies persist, leading to increased costs and time-to-market.

Simulation-based verification, the industry standard, offers high coverage but is computationally expensive and time-consuming. As \ac{IC} complexity scales, exhaustive state-space validation becomes impractical due to exponential verification effort. Formal verification, while mathematically rigorous, suffers from scalability limitations arising from manual property specification, model complexity, and challenges in verifying deeply sequential designs. These constraints have driven the search for AI-driven verification methodologies that enhance efficiency and throughput while reducing manual effort.

Recent advances in \ac{GenAI} and \acp{LLM} have demonstrated potential in \ac{IC} design and verification \cite{unknown}, enabling tasks such as \ac{RTL} generation \cite{islam2024aivrilaidrivenrtlgeneration}, testbench creation \cite{huang2024llmpoweredverilogrtlassistant}, stimuli generation \cite{zhang2023llm4dvusinglargelanguage}, coverage closure \cite{10546707}, formal property generation \cite{fang2024assertllmgeneratingevaluatinghardware}, and vulnerability detection \cite{akyash2024selfhwdebugautomationllmselfinstructing}. Frameworks like AIVRIL \cite{islam2024aivrilaidrivenrtlgeneration} and VeriAssist \cite{huang2024llmpoweredverilogrtlassistant} leverage \acp{LLM} for \ac{RTL} design but face scalability challenges due to their dependence on simulation. Moreover, applying \acp{LLM} to hardware design poses significant difficulties, as studies \cite{gadde2024artificialintelligencegenailens} report around \SI{60}{\percent} failure rates in generated \ac{RTL} designs due to issues like randomness, hallucinations, and challenges in handling complex specifications. These findings highlight the overly optimistic expectations of zero-shot \ac{LLM} applications in hardware design tasks. Moreover, existing approaches often focus on static code generation but overlook iterative refinement, self-correction, and \ac{HITL} integration, which are essential for reliable design and verification. To address these limitations, this work proposes a methodology based on Agentic AI, leveraging a \ac{MAS} \cite{chen2025surveyllmbasedmultiagentsystem} to integrate autonomous AI agents for \ac{RTL} generation and formal verification. Through \ac{MAS}-driven adaptability, agentic coordination, and \ac{HITL}-enhanced verification, this approach establishes a scalable, intelligent, and high-confidence verification paradigm for next-generation \ac{IC} design.

The key contributions of this work are sketched as follows:
\begin{itemize}
	\item Unlike conventional \ac{LLM}-based approaches such as zero-shot, our work proposes \ac{MAS} which enables collaborative agent-based code generation, iterative \ac{RTL} refinement, dynamic property generation 
	
	\item Fully autonomous end-to-end workflow where specialized agents can not only generate code but also interact directly with industry-standard EDA tools 
	
	\item \ac{HITL} integration to resolve ambiguities in AI-generated \ac{RTL} and formal properties 
\end{itemize}

\section{Agentic Workflows} \label{sec:background}
\begin{figure}[htbp]
	\centering
	\includegraphics[width=\linewidth, keepaspectratio]{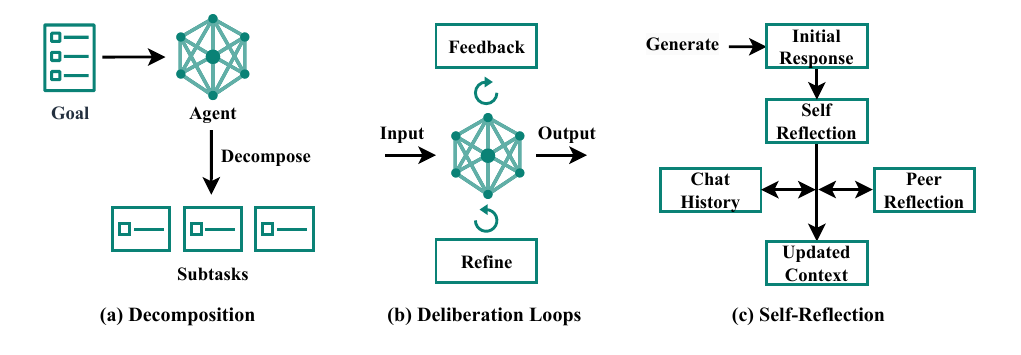}
	\caption{Agentic design patterns}
	\label{agentic_design_patterns}
\end{figure}

Traditional \ac{RTL} design and verification workflows are predominantly non-agentic, relying on sequential, manual processes automated by scripting and EDA tools. In non-agentic workflows, engineers manually write \ac{RTL} code or use template-based code generators, followed by the separate development of testbenches and formal properties. In contrast, agentic workflows distribute tasks among autonomous AI agents that coordinate to achieve a common goal. Rather than a single \ac{LLM} executing prompts in isolation, multiple specialized agents each equipped with distinct capabilities collaborate through structured interaction patterns, as shown in Fig. \ref{agentic_design_patterns}. The most prominent agentic design patterns include:
\begin{enumerate}[label=(\alph*)]
	\item Pipeline decomposition, which involves breaking down larger tasks into smaller actionable subtasks that agents can execute independently \cite{jeyakumar2024advancing}
	\item Deliberation loops, in which agents iteratively propose designs or properties and critique one another to converge on high‑quality outputs making final output more robust \cite{chen2025surveyllmbasedmultiagentsystem}
	\item Reflection and self‑correction, enabling agents to review previous outputs, identify errors, and autonomously initiate re‑generation enhancing reliability \cite{10852426}
\end{enumerate}

By enabling agents to think, research, generate, and review designs before producing final outputs, agentic workflows significantly improve reliability compared to zero-shot approaches while addressing key challenges including explainability of decisions, error propagation prevention, seamless EDA tool integration, appropriate human oversight, and coding standards compliance \cite{patra2024aiedaagenticaidesign}. These workflows are orchestrated via multi-agent collaboration topologies ranging from broadcast group chats to hierarchical or role-based channels that assign tasks to specialized sub-agents, enabling dynamic information routing, peer review among agents, and efficient coordination that improves scalability, reduces redundancy, and fosters specialized expertise while maintaining \ac{HITL} intervention for critical decision points.

\section{Methodology} \label{sec:methodology}
\begin{figure}[htbp!]
	\centering
	\includegraphics[width=0.9\linewidth, keepaspectratio]{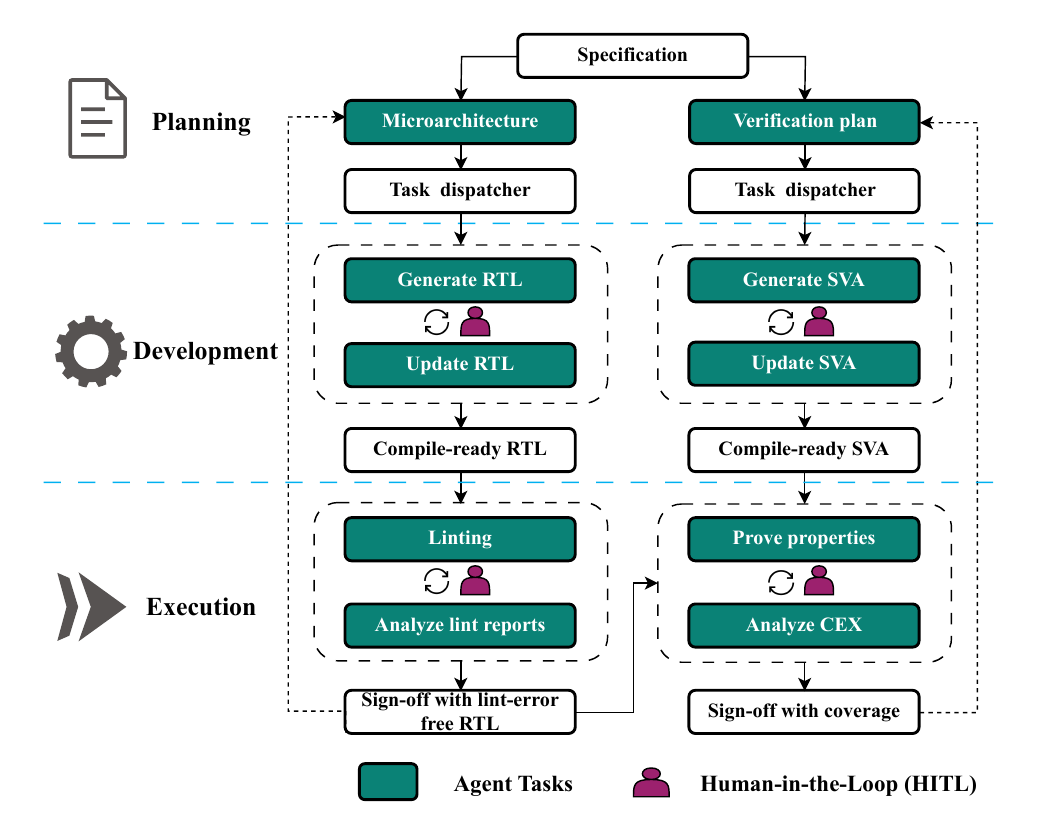}
	\caption{Agentic AI methodology for RTL design and verification with HITL}
	\label{agentic_flow}
\end{figure}

The proposed methodology leverages MAS paradigm to automate \ac{RTL} design and verification flow integrating \ac{HITL} interventions to ensure reliability and compliance. This approach is illustrated in Fig. \ref{agentic_flow}. At its core, the system employs specialized AI agents that collaborate to transform high‑level specifications into verified \ac{RTL} implementations, balancing autonomous generation with human guidance. The process is structured into three distinct phases: planning, development, and execution. This structure ensures alignment with original requirements while facilitating iterative feedback and escalation when necessary.

The process begins with a structured specification document encompassing natural‑language requirement descriptions, interface definitions, performance targets, and FSM details. In the planning phase, the \textit{design lead} agent parses the specification to produce a microarchitecture detailing datapath components, control‑state machines, reset strategies, timing constraints and more in a structured format. Concurrently, the \textit{formal verification lead} agent interprets the same specification to generate a verification plan (vPlan), outlining property types, coverage goals and more. By deriving both the flows from a unified specification, any divergence between design intent and verification objectives are eliminated from the beginning.

During the development phase, task dispatchers channel the outputs from the planning phase into two parallel streams. In the design stream, \ac{RTL} agents utilize the microarchitectural guidelines to generate SystemVerilog modules for each functional block, complying to coding standards and macro templates provided. In the verification stream, formal verification agents translate each vPlan entry into \acp{SVA}. Throughout this phase, critic agents monitor outputs flagging issues such as missing logic, syntax errors, coverage gaps and more. If predefined iteration limits are reached without resolution, a \textit{conversable agent} escalates the issue to human reviewers for \ac{HITL} intervention, addressing ambiguities beyond the capabilities of automated agents.

In the execution phase, the code executor agent interfaces with industry-standard EDA tools to validate the generated \ac{RTL} and \acp{SVA}. A dedicated \textit{code executor agent} invokes linting tools such as SpyGlass \cite{synopsysSpyGlassLint} to ensure no lint errors are present. Lint errors are parsed, categorized, and returned to the \ac{RTL} agents for automated patching or routed to \ac{HITL} when errors exceed the agents repair capabilities. For formal verification, the \textit{code executor agent} drives formal tools such as JasperGold \cite{cadenceJasper} to prove each SVA. \acp{CEX} are analyzed, prompting iterative fixes by development agents or escalation to human engineers for resolution. A dedicated coverage agent consolidates code coverage, assertion coverage, and functional coverage metrics against the original vPlan targets. If coverage thresholds are unmet, the system iteratively generates additional properties to achieve defined goals. The methodology concludes with a formal sign‑off, delivering a compile‑ready \ac{RTL} along with a structured coverage report detailing proven properties, coverage results, and any exceptions.

\subsection{Multi-Agent Framework Architecture}

The proposed methodology employs a modular agent orchestration system designed to be compatible with open-source \ac{MAS} including CrewAI \cite{crewai2024}, AutoGen \cite{autogen2024}, and LangGraph \cite{langgraph2024} while the current evaluation focus exclusively on Autogen. It facilitates scalable \ac{MAS} through event-driven model enabling deterministic agent interactions which manages agent roles, responsibilities, and interaction patterns, enabling dynamic adaptation to diverse design requirements. The framework supports seamless integration of various \acp{LLM} including GPT-4o \cite{openai2025chatgpt4o} and Llama3.1 \cite{llama3meta} through a standardized API layer. This modularity allows for hot-swapping of additional models, agents, tools with minimal code modifications ensuring adaptability to evolving design requirements. Each agent within the system is assigned with specific roles which then coordinates with other agents through structured protocols.

\begin{figure}[ht]
	\vspace*{-0.5cm}
	\centering
	\includegraphics[width=\linewidth, keepaspectratio]{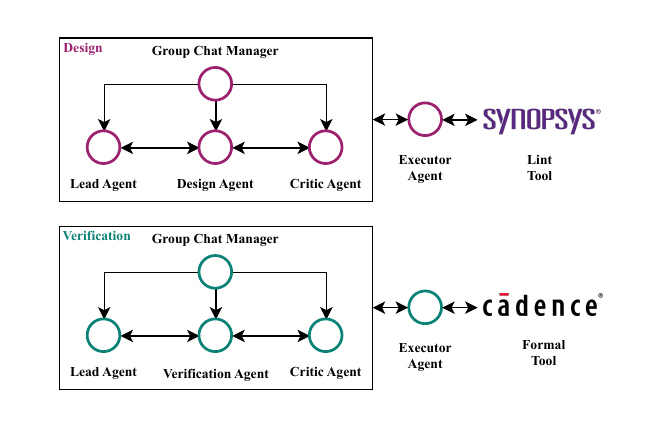}
	\vspace*{-0.8cm}
	\caption{Illustration of agentic AI workflow execution}
	\label{agent_orchestration}
\end{figure}

Within the framework, AI agents collaborate to execute tasks specified by human users within a multi-agent group chat environment as shown in Fig. \ref{agent_orchestration}. Each agent specializes in specific roles such as \textit{verification lead}, \textit{formal verification engineer}, and \textit{SystemVerilog LRM expert}. This design pattern enables agents to share a common message thread by subscribing and publishing to the same topic. Additionally, a \ac{HITL} agent represents the human user, providing oversight and guidance when necessary to ensure the verification process aligns with human expertise and decision-making. The \textit{Group Chat Manager} agent orchestrates the sequential interaction, ensuring that only one agent communicates at a time using AutoGen’s Core API and an event-driven architecture.

To address common limitations of \acp{LLM} such as attention deficits, hallucination, and getting stuck in iterative loops, the methodology decomposes tasks into smaller tasks distributed among multiple agents. Throughout this process, critic agents evaluate the generated properties, providing feedback to enhance the quality and correctness of generated \acp{SVA}. A feedback loop with a threshold of five iterations is established; exceeding this limit triggers human intervention to prevent workflow stagnation and reduce hallucination errors.

The system incorporates comprehensive logging mechanisms that capture detailed interactions at multiple granularity levels including agent group chats, tool integration mechanisms, and error management systems. An integrated exception management system and threshold-based monitoring prevent infinite loops, automatically triggering human intervention when autonomous resolution is unattainable. Through this architecture, the methodology establishes a scalable, intelligent, and high-confidence verification paradigm for next-generation \ac{IC} design, leveraging the strengths of both automated AI agents and strategic human oversight.

\section{Evaluation} \label{sec:results}
The effectiveness of the proposed \ac{MAS} methodology is evaluated by applying it to a range of internal and open-source design specifications. For benchmarking, five open-source designs—CRC, ECC, FIFO, Lemming, and Timer—were utilized. The specifications for CRC, ECC, and FIFO were sourced from OpenCores \cite{opencores}, while those for Lemming and Timer were obtained from VerilogEval \cite{pinckney2024revisitingverilogevalnewerllms}. The framework employs OpenAI's GPT-4o model, a 1.8 trillion parameter \ac{LLM}, to generate and verify the \ac{RTL}. To evaluate hyperparameter effects, the methodology was run at \textit{\ac{temp}} of 0.2, 0.5, and 0.8, where lower values yield consistent results and higher values foster diverse output.

The \ac{RTL} generation phase was evaluated using quality metrics, including lint error counts, logical correctness (classified as correct, correct but non-synthesizable, incorrect, or incomplete), and the time required for \ac{HITL} intervention to achieve reliable \ac{RTL}. For the formal verification phase, metrics such as the number of generated properties, achieved formal coverage percentage, \ac{CEX} count, and additional \ac{HITL} effort needed to approach near \SI{100}{\percent} coverage were assessed. The \ac{HITL} property count represents the final set of properties after redundancies are removed and new properties are added. The comprehensive results, presented in Table~\ref{tab:results_summary}, demonstrate the framework's effectiveness across designs of varying complexity, providing insights into the relationship between \textit{\ac{temp}} settings and design quality.

\begin{table}[htbp!]
	\centering
	\caption{Overview of the results}
	\label{tab:results_summary}
	\begin{threeparttable}
		\adjustbox{width=\linewidth}{
			\begin{tabular}{|c|c|c|c|c|c|c|c|c|c|c|c|}
				\hline
				\multirow{3}{*}{\textbf{Design}} & \multirow{3}{*}{\textbf{Temp}} & \multicolumn{4}{c|}{\textbf{RTL Generation}} & \multicolumn{6}{c|}{\textbf{Formal Property Generation}} \\ \cline{3-12} 
				&  & \multicolumn{2}{c|}{\textbf{\#Errors (linting)}} & \multicolumn{2}{c|}{\textbf{Logical Accuracy}} & \multicolumn{2}{c|}{\textbf{\#Properties}} & \multicolumn{2}{c|}{\textbf{Coverage (\%)}} & \multicolumn{2}{c|}{\textbf{\#CEXs}} \\ \cline{3-12}
				&  & \textbf{MAS} & \textbf{HITL} & \textbf{MAS} & \textbf{HITL} & \textbf{MAS} & \textbf{HITL} & \textbf{MAS} & \textbf{HITL} & \textbf{MAS} & \textbf{HITL} \\ \hline
				\multirow{3}{*}{CRC\cite{opencores}} & 0.2 & 0 & 0 & \faCheck & \faCheck & \multirow{3}{*}{16} & \multirow{3}{*}{11} & \multirow{3}{*}{73.08} & \multirow{3}{*}{100} & \multirow{3}{*}{8} & \multirow{3}{*}{0} \\ \cline{2-6}
				& 0.5 & 0 & 0 & \faExclamation & \faCheck &  &  &  &  &  &  \\ \cline{2-6}
				& 0.8 & 0 & 0 & \faTimes & \faCheck &  &  &  &  &  &  \\ \hline
				\multirow{3}{*}{ECC\cite{opencores}} & 0.2 & 5 & 0 & \faBan & \faCheck & \multirow{3}{*}{19} & \multirow{3}{*}{12} & \multirow{3}{*}{93.88} & \multirow{3}{*}{95.90*} & \multirow{3}{*}{8} & \multirow{3}{*}{0} \\ \cline{2-6}
				& 0.5 & 2 \faFlash & 0 & \faBan & \faCheck &  &  &  &  &  &  \\ \cline{2-6}
				& 0.8 & 2 \faFlash & 0 & \faBan & \faCheck &  &  &  &  &  &  \\ \hline
				\multirow{3}{*}{FIFO\cite{opencores}} & 0.2 & 0 & 0 & \faCheck & \faCheck & \multirow{3}{*}{20} & \multirow{3}{*}{13} & \multirow{3}{*}{91.67} & \multirow{3}{*}{97.29*} & \multirow{3}{*}{11} & \multirow{3}{*}{0} \\ \cline{2-6}
				& 0.5 & 0 & 0 & \faTimes & \faCheck &  &  &  &  &  &  \\ \cline{2-6}
				& 0.8 & 0 & 0 & \faCheck & \faCheck &  &  &  &  &  &  \\ \hline
				\multirow{3}{*}{Lemming\cite{pinckney2024revisitingverilogevalnewerllms}} & 0.2 & 0 & 0 & \faTimes & \faCheck & \multirow{3}{*}{18} & \multirow{3}{*}{21} & \multirow{3}{*}{96.15} & \multirow{3}{*}{96.77*} & \multirow{3}{*}{3} & \multirow{3}{*}{0} \\ \cline{2-6}
				& 0.5 & 0 & 0 & \faTimes & \faCheck &  &  &  &  &  &  \\ \cline{2-6}
				& 0.8 & 1 \faFlash & 0 & \faExclamation & \faCheck &  &  &  &  &  &  \\ \hline
				\multirow{3}{*}{Timer\cite{pinckney2024revisitingverilogevalnewerllms}} & 0.2 & 0 & 0 & \faExclamation & \faCheck & \multirow{3}{*}{42} & \multirow{3}{*}{57} & \multirow{3}{*}{83.95} & \multirow{3}{*}{96.39*} & \multirow{3}{*}{22} & \multirow{3}{*}{0} \\ \cline{2-6}
				& 0.5 & 0 & 0 & \faTimes & \faCheck &  &  &  &  &  &  \\ \cline{2-6}
				& 0.8 & 0 & 0 & \faTimes & \faCheck &  &  &  &  &  &  \\ \hline
			\end{tabular}
		}
		\begin{center}
			\vspace{1ex}
			\scriptsize 
			\faFlash: Fatal, \faCheck: Correct, \faExclamation: Non-synthesizable, \faTimes: Incorrect, \faBan: Incomplete. \\
			\vspace{1ex} 
			\parbox{\columnwidth}{
				\scriptsize{The \ac{HITL} effort for \ac{RTL} and Formal Property Generation was 15/40 minutes for CRC, 30/20 minutes for ECC, 15/20 minutes for FIFO, 15/15 minutes for Lemming and 15/30 minutes for Timer, respectively. \\
					* Indicates the uncovered portion is unreachable dead code}
			}
		\end{center}
	\end{threeparttable}
\end{table}

\subsection{\ac{RTL} Generation}
The evaluation demonstrated that generated \ac{RTL} for single module designs like CRC showed no lint errors across \textit{\ac{temp}}. Higher \textit{\ac{temp}} introduced randomness, occasionally producing incorrect \ac{RTL}, as shown in Table~\ref{tab:results_summary}. Brief \ac{HITL} interventions, typically under {\SI{15}{\minute}, refined the \ac{RTL} for better usability. For ECC design at 0.3 \textit{\ac{temp}}, two modules emerged with functional placeholders, causing lint errors. Specification refinement and adjusted agent configurations resolved these issues. Subsequent \ac{HITL} efforts under {\SI{30}{\minute} completed missing logic. Higher \textit{\ac{temp}} produced severe lint errors, promptly addressed through targeted \ac{HITL}. FIFO exhibited lint-free \ac{RTL} across varying \textit{\ac{temp}}, demonstrating robust quality. At \textit{\ac{temp}} 0.5, a minor logical error was resolved with {\SI{15}{\minute} of \ac{HITL} effort.
Lemming passed linting for all \textit{\ac{temp}} values except 0.8, where a syntax error was identified. Functional bugs, common across all \textit{\ac{temp}} values, were resolved with \ac{HITL} efforts, making the \ac{RTL} ready for formal verification. Similarly, Timer designs passed linting at all \textit{\ac{temp}} values, but control logic bugs were present across all cases and were addressed through targeted \ac{HITL} intervention.
			
The \ac{RTL} generation process required varying numbers of iterative refinement cycles depending on design complexity. The Timer design converged after just 1 iteration, while CRC, FIFO, and Lemming designs required 2 iterations. The more complex ECC design, however, needed 3 iterations to achieve stable, functional \ac{RTL}. The nature of human interventions during \ac{RTL} generation primarily involved removing placeholder code segments and providing design-specific feedback to guide the agents. This iterative feedback loop improved the accuracy of the generated \ac{RTL} while minimizing the need for extensive human rewrites.

\subsection{Formal Property Generation}
Following the lint-error free \ac{RTL} generation, the verification flow generates formal properties, which are then validated and analyzed for coverage using formal tool, as detailed in Table~\ref{tab:results_summary}. The \ac{MAS} approach achieved an average initial coverage of \SI{86.21}{\percent} across all designs, with generated properties demonstrating high compilation success rates. Strategic \ac{HITL} interventions, averaging \SI{27}{\minute} per design, refined the properties to achieve \SI{100}{\percent} assertion pass rate and an average final coverage of \SI{97.73}{\percent}, with remaining uncovered portions primarily consisting of unreachable dead code and default case statements.

The comparative analysis presented in Fig.~\ref{fig:results} demonstrates that while zero-shot approaches achieved modest initial coverage averaging \SI{69.85}{\percent}, they required extensive prompt adjustments and iterative refinements even to ensure property compilation. In contrast, the proposed \ac{MAS} approach demonstrated superior baseline performance with minimal iterations and, with \ac{HITL} integration, achieved \SI{100}{\percent} assertional pass rate and nearly \SI{98}{\percent} coverage, demonstrating the effectiveness of the collaborative agent-human verification paradigm.

The human interventions during property generation focused primarily on addressing conceptual issues through targeted feedback and removing redundant properties. This complementary human-AI workflow capitalized on the agents' comprehensive property generation capabilities while leveraging human expertise to optimize verification coverage, resulting in higher quality properties and more efficient verification processes.

\subsection{Discussion}

The proposed methodology effectively manages complex design and verification tasks through specialized agents and process decomposition, addressing common \ac{LLM} limitations. In contrast to zero-shot approach, \ac{MAS} facilitates dynamic reasoning and iterative refinement where \ac{HITL} interventions enhance robustness by resolving \ac{LLM} errors and incomplete outputs, enabling accurate and complete \ac{RTL} generation and verification through human-AI collaboration. The effectiveness of code generation heavily depends on the underlying quality of \acp{LLM} used, precise agent configurations through prompt engineering, and careful tuning of hyper parameters. While critic agents ensure the quality and correctness of generated properties through continuous feedback, their performance is contingent upon optimal agent count and configuration. Nevertheless, when properly configured, the framework's systematic task division, agent specialization, and \ac{HITL} ensure comprehensive coverage, error-free \ac{RTL} generation, and reliable verification.

\begin{figure}[htpb!]
	\vspace*{-0.2cm}
	\centering
	\includegraphics[width=0.9\linewidth]{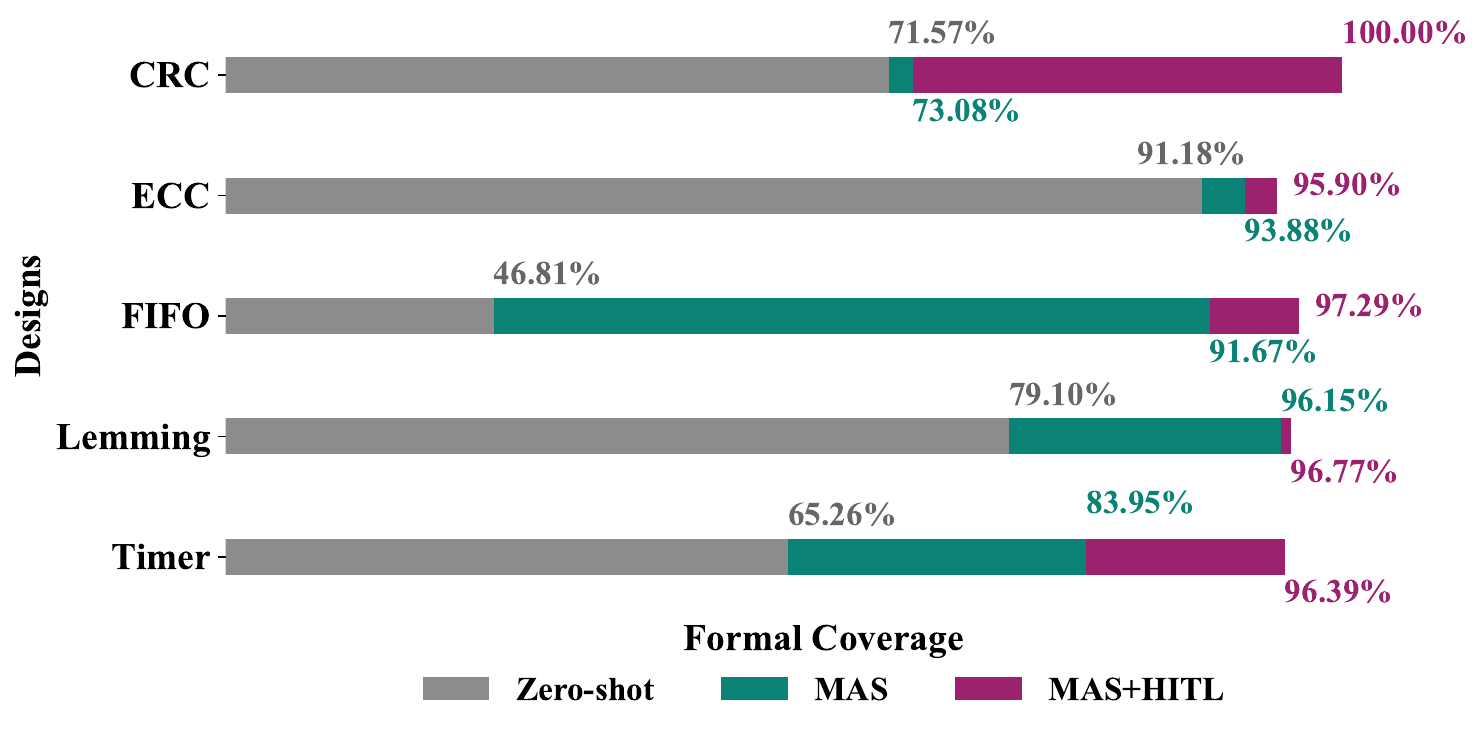}
	\caption{Comparison of coverage results produced by proposed and zero-shot approaches}
	\label{fig:results}
\end{figure}

The \ac{HITL} intervention model introduces an interesting cognitive trade-off. While human intervention times were relatively short (under 40 minutes), the cognitive effort required to analyze AI-generated designs may differ from reviewing human-created designs. The intervention process requires engineers to understand not only the design intent but also the reasoning patterns and potential limitations of the AI agents. However, the targeted nature of the interventions—focused on specific placeholders, lint errors, or counterexamples—reduces the cognitive load compared to comprehensive manual reviews.

\section{Conclusion} \label{sec:conclusion}

This paper presents a novel \ac{MAS}-based methodology for automated RTL design and verification, achieving around \SI{95}{\percent} coverage through collaborative specialized agents and targeted \ac{HITL} interventions. The approach demonstrates superior performance compared to zero-shot methods across various designs, with \ac{MAS}+\ac{HITL} achieving up to \SI{100}{\percent} coverage while requiring minimal human effort. Although not guaranteeing \SI{100}{\percent} efficacy in every run, the methodology consistently demonstrates high-quality results, with robust performance influenced by hyper parameter settings and agent configurations. The hot-pluggable domain-agnostic architecture facilitates integration of emerging models, enabling adaptation to advances in \ac{LLM} models, while the scalability analysis suggests applicability to more complex designs. Future work would focus on enhanced agent coordination, deeper EDA tool integration, explainable AI techniques to reduce human intervention advancing the state of AI-driven hardware design and verification towards truly autonomous systems. Furthermore, the methodology is being expanded to generate efficient testbenches and stimuli that addresses the verification of more intricate designs.

\printbibliography

\end{document}